\documentclass[lettersize,journal]{IEEEtran}
\usepackage{cite}
\usepackage{amsmath,amssymb,amsfonts}
\usepackage{graphicx}
\usepackage{textcomp}

\usepackage{algorithmic}
\usepackage{algorithm}
\usepackage{array}
\usepackage[caption=false,font=normalsize,labelfont=sf,textfont=sf]{subfig}
\usepackage{textcomp}
\usepackage{stfloats}
\usepackage{url}
\usepackage{verbatim}
\usepackage{graphicx}
\usepackage{cite}
\usepackage{graphicx}
\usepackage{amssymb}
\usepackage{hyperref}
\usepackage{dsfont}
\usepackage{epstopdf}

\usepackage{amsmath}
\usepackage{multirow}
\usepackage{soul}
\usepackage{booktabs}
\usepackage{bbding}
\usepackage{makecell}
\usepackage{stfloats}
\usepackage{xcolor}
\usepackage{makecell}
\usepackage{xspace}
\usepackage{url}
\usepackage{wrapfig}
\usepackage{multirow}
\usepackage{makecell}
\usepackage{booktabs, colortbl}
\usepackage{longtable}
\usepackage{hhline}

\definecolor{myblue}{HTML}{0072C6}
\definecolor{myyellow}{HTML}{FFFADF}
\definecolor{myred}{HTML}{FF0000}

\def\ie{\emph{i.e.}}
\def\eg{\emph{e.g.}}
\def\etal{{\em et al.}}
 
\usepackage{rotating}

\def\BibTeX{{\rm B\kern-.05em{\sc i\kern-.025em b}\kern-.08em
    T\kern-.1667em\lower.7ex\hbox{E}\kern-.125emX}}
\begin{document}
\title{Vivim: a Video Vision Mamba for\\ Medical Video Segmentation}
\author{Yijun Yang,
        Zhaohu Xing,
        Lequan Yu,~\IEEEmembership{Member,~IEEE,}
	Chunwang Huang,
	Huazhu Fu,~\IEEEmembership{Senior Member,~IEEE,}
	Lei Zhu,~\IEEEmembership{Member,~IEEE}
	\thanks{Manuscript  received  XX,  2024;  revised XX,  2024; accepted XX, 2024. 
	}
	\thanks{Yijun Yang and Zhaohu Xing are with Robotics and Autonomous Systems Thrust, The Hong Kong University of Science and Technology, Guangzhou, China (e-mail: yyang018@connect.hkust-gz.edu.cn; zxing565@connect.hkust-gz.edu.cn).}
	\thanks{Lequan Yu is with the Department of Statistics and Actuarial Science, The University of Hong Kong, Hong Kong SAR, China (e-mail: lqyu@hku.hk).}
	\thanks{Chunwang Huang is with Guangdong Provincial People's Hospital, Guangzhou, China (e-mail: huangchunwang@126.com).}
        \thanks{Huazhu Fu is with the Institute of High Performance Computing, A*STAR, Singapore (e-mail: hzfu@ieee.org).} 
        \thanks{Lei Zhu is with Robotics and Autonomous Systems Thrust, Hong Kong University of Science and Technology (Guangzhou), China, and the Department of Electronic and Computer Engineering, Hong Kong University of Science and Technology, Hong Kong SAR, China (e-mail: leizhu@ust.hk).}
	\thanks{Lei Zhu is the corresponding author of this work.}
}

\maketitle

\begin{abstract}
Medical video segmentation gains increasing attention in clinical practice due to the redundant dynamic references in video frames.
However, traditional convolutional neural networks have a limited receptive field and transformer-based networks are mediocre in constructing long-term dependency from the perspective of computational complexity.
This bottleneck poses a significant challenge when processing longer sequences in medical video analysis tasks using available devices with limited memory.
Recently, state space models (SSMs), famous by Mamba, have exhibited impressive achievements in efficient long sequence modeling, which develops deep neural networks by expanding the receptive field on many vision tasks significantly.
Unfortunately, vanilla SSMs failed to simultaneously capture causal temporal cues and preserve non-casual spatial information.
To this end, this paper presents a Video Vision Mamba-based framework, dubbed as \textbf{Vivim}, for medical video segmentation tasks.
Our Vivim can effectively compress the long-term spatiotemporal representation into sequences at varying scales with our designed Temporal Mamba Block.
We also introduce an improved boundary-aware affine constraint across frames to enhance the discriminative ability of Vivim on ambiguous lesions.
Extensive experiments on thyroid segmentation, breast lesion segmentation in ultrasound videos, and polyp segmentation in colonoscopy videos demonstrate the effectiveness and efficiency of our Vivim, superior to existing methods.
The code is available at:
\href{https://github.com/scott-yjyang/Vivim}{https://github.com/scott-yjyang/Vivim}.
The dataset will be released once accepted.
\end{abstract}


\section{Introduction}

Automatic segmentation of lesions and tissues is essential for computer-aided clinical examination and treatment~\cite{huang2020segmentation}, such as ultrasound lesion segmentation, polyp segmentation.
%
% However, lesions are usually situated in complex, dynamic environments where surrounding tissues or fluids are constantly moving or changing. 
However, segmenting medical objects is usually challenging due to inherent factors, including ambiguous lesion boundaries, inhomogeneous distributions, diverse motion patterns, and dynamic changes in complex environments~\cite{lin2022new}.
Medical videos, essentially sequences of medical images, offer a richer and more detailed context for locating ambiguous lesions and tissues. 
This additional information makes video-based segmentation better handle unexpected complexities by providing a continuous view, allowing for a more accurate and comprehensive analysis.
Consequently, to consider more object context, expanding the deep model's receptive field in the spatiotemporal space is highly desired in medical video analysis. 
Traditional convolutional neural networks~\cite{zhou2018unet++,he2016deep,he2017mask,yang2021hcdg} often struggle to capture global information compared to recent transformer-based architectures.
The transformer architecture, which utilizes the Multi-Head Self Attention (MSA)~\cite{vaswani2017attention} to extract global information, has attracted much attention from the community of generic video object segmentation~\cite{zheng2021rethinking,arnab2021vivit,liang2020video}.
Considering that neighboring frames offer beneficial hints to the segmentation, these methods usually introduce some elaborated modules on the self-attention mechanism to exploit the temporal information.
However, exploring the additional temporal dimension often leads to increased complexity and greater demands on resources, posing significant challenges for implementation due to the strict environmental conditions and inherently high-dimensional characteristics of medical videos. 
For example, the incorporation of temporal self-attention modules can unintentionally trigger a quadratic increase in complexity relative to the time dimension, resulting in substantial computational challenges. 
The marked rise in the number of tokens within lengthy video sequences introduces considerable computational strains when employing Multi-head Self-Attention (MSA) techniques for temporal information modeling~\cite{arnab2021vivit}.

Very recently, to address the ill-posed issue concerning long sequence modeling, Mamba~\cite{gu2023mamba}, inspired by state space models (SSMs)~\cite{kalman1960new}, has been developed. Its main idea is to efficiently capture long-range dependencies by implementing a selective scan mechanism for 1-D sequence interaction. 
Based on this, U-Mamba~\cite{ma2024u} designed a hybrid CNN-SSM block, which is mainly composed of Mamba modules, to handle the long sequences in biomedical image segmentation tasks.
Vision Mamba~\cite{zhu2024vision} provided a new generic vision backbone with bidirectional Mamba blocks on image classification and semantic segmentation tasks.
As suggested by them, relying on the self-attention module is not necessary to achieve efficient visual representation learning. 
It can be replaced by Mamba when exploring long-term temporal dependency in video scenarios. 
The crucial aspect of adapting the Vision Mamba model for video applications lies in the ability to concurrently capture causal temporal cues while maintaining the integrity of non-causal spatial information.
% Inherently, this offers an effective solution to explore long-term temporal dependency in video scenarios.
%

\begin{figure*}[!t]
\centering
\includegraphics[width=\textwidth]{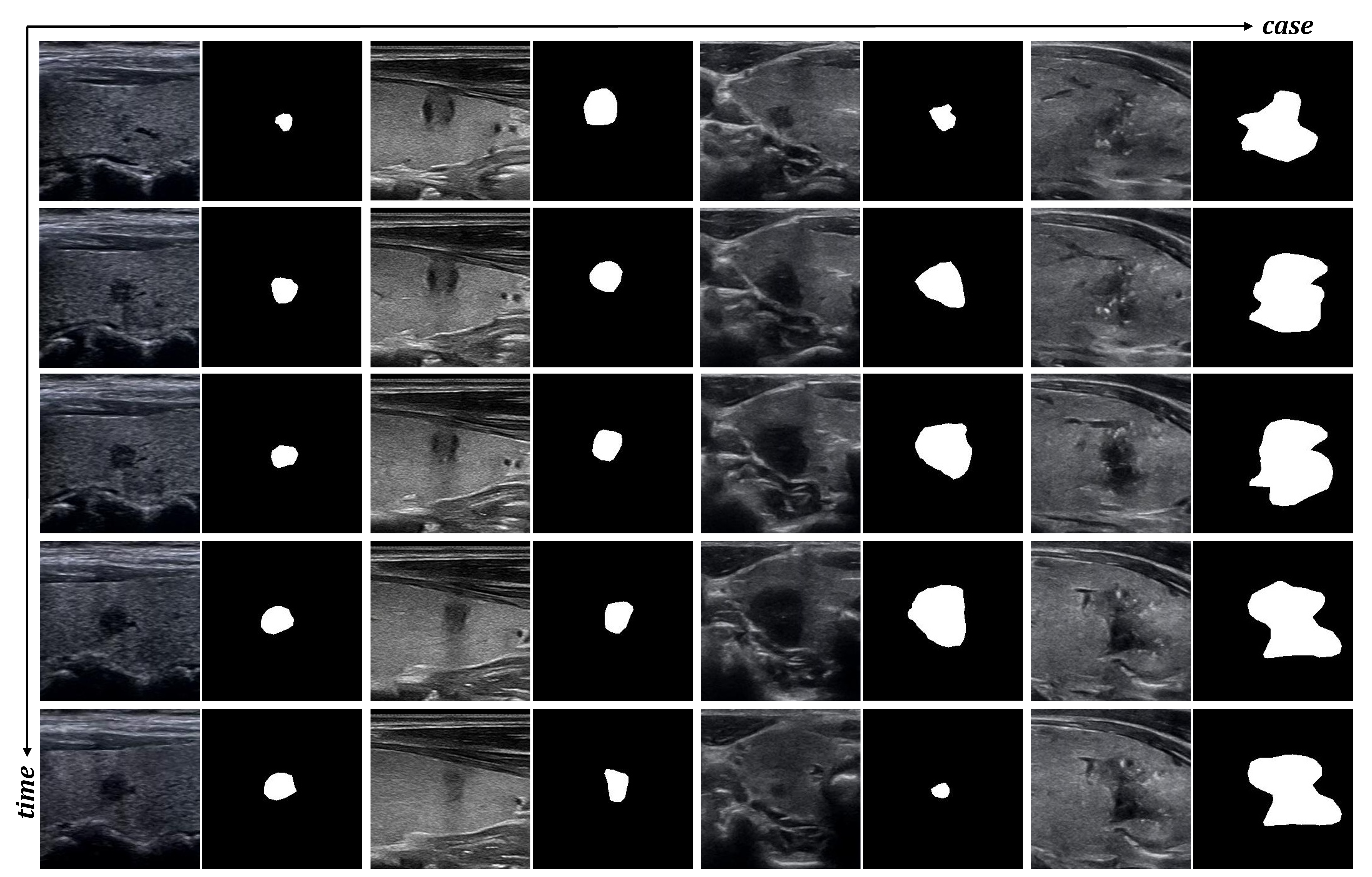}
\caption{
\textbf{Several cases of our collected VTUS dataset.} All videos are taken from patients with thyroid nodules. They are taken by ultrasound doctors with more than 10 years of clinical experience to ensure the image quality. These videos are cross-annotated by three experts with over three years of experience in thyroid diagnosis.
} 
\label{fig:vtus_data}
\end{figure*}

Motivated by this, we present an SSMs-based framework Vivim that integrates Mamba into the multi-level transformer architecture to exploit spatiotemporal information in videos with linear complexity.
% to construct the feature sequence containing spatiotemporal information from videos. 
% Our Vivim is designed to explore the temporal dependency across frames for the advancement of segmentation results with the cheaper computational cost than other video-based methods.
%
\textit{To the best of our knowledge, this is the first work to incorporate SSMs into the task of medical video segmentation, facilitating faster and greater performance.}
In our Vivim, drawing inspiration from the architecture of modern transformer blocks, we design a novel Temporal Mamba Block. 
A hierarchical encoder consisting of multiple Temporal Mamba Blocks is introduced to investigate the correlation between spatial and temporal dependency at various scales.
As the structured state space sequence models with selective scan (S6)~\cite{gu2023mamba} causally process input data, they can only capture information within the scanned portion of the data. This aligns S6 with NLP and video tasks involving temporal data but poses challenges when addressing non-causal data like 2D images within medical videos. 
To this end, the structured state space sequence model with spatiotemporal selective scan, ST-Mamba, is designed and incorporated into each scale of the model's encoder, replacing the self-attention or window-attention module to achieve efficient video visual representation learning.
Finally, we employ an improved boundary-aware affine constraint to improve the discrimination of Vivim on ambiguous tissues in medical videos at the training stage.

It is worth noting that \textit{there is no public dataset with pixel-level annotated ultrasound videos for thyroid segmentation}, as it is expensive to delineate the boundaries of ambiguous lesions in low-contrast ultrasound videos in a frame-by-frame spirit. 
In this work, we collect a thyroid segmentation dataset VTUS with 100 annotated transverse viewed and longitudinal viewed ultrasound videos and a total of 9342 frames with pixel-level ground truth to facilitate the benchmarking evaluation. Several examples are displayed in Fig. \ref{fig:vtus_data}.
We conduct extensive experiments on three popular medical video segmentation tasks, \ie, thyroid segmentation in ultrasound videos, breast lesion segmentation in ultrasound videos, and polyp segmentation in colonoscopy videos. The superior results validate the effectiveness, efficiency and versatility of our framework Vivim.
% Experiments on breast US videos demonstrate the effectiveness of our framework Vivim.

Our contributions can be summarized as follows:
\begin{itemize}

    \item We develop a medical video segmentation framework consisting of a Mamba-based encoder and a CNN-based decoder to obtain holistic understanding of medical videos and preserve local details, respectively. This is the first work to introduce state space models into medical video scenarios.
    \item Instead of simply adapting Mamba to medical tasks, we design spatio-temporal selective scan to enhance the global perception capability in videos of our Temporal Mamba Block.
    \item We employ an improved boundary-aware constraint based on the optimization of the affine transformation to mitigate ambiguous boundary prediction of our model.
    \item We collect the first video ultrasound thyroid segmentation dataset with pixel-level annotation, which facilitates the benchmarking evaluation of medical video segmentation methods. Our model achieves promising segmentation results on diverse modalities but maintains decent efficiency superior to Transformer-based methods.

\end{itemize}

\section{Related Works}

% \subsection{Thyroid Segmentation}

\subsection{Medical Video Segmentation}

Recent approaches have introduced innovative hybrid transformer-based algorithms that fuse transformative and convolutional layer techniques for medical image segmentation (\eg, breast lesion, polyp)~\cite{he2023transformers,he2023swinunetr,zhou2018unet++,hatamizadeh2022unetr,chen2021transunet,wang2018deep,ali2024objective,yang2021hcdg}. 
For thyroid segmentation in ultrasound images, Jeremy \etal~\cite{webb2020automatic} developed a novel spatio-temporal recurrent deep learning network to automatically segment the thyroid gland in ultrasound cineclips by leveraging time sequence information. 
Ma \etal~\cite{ma2022novel} utilized the region proposal network (RPN) for initial deep feature extraction and incorporated the spatial pyramid RoIAlign as a segmentation head to capture global and local information in ultrasound images.
Chi \etal~\cite{chi2023hybrid} developed a 2D Transformer-UNet for thyroid gland segmentation, combining high-level features from decoding layers with lower-level features from encoding layers using a multiscale cross-attention transformer module.
These algorithms skillfully manage the representations derived from high-definition medical images, however, they grapple with computational difficulties owing to complexity issues. 
Additionally, the direct application of such image segmentation methods may inadvertently overlook critical temporal context, thereby inducing temporal inconsistencies. 
In order to address temporal modeling in video-level segmentation, the innovative method of Space-Time Memory Networks (STM)~\cite{oh2019video} and its variants~\cite{cheng2021rethinking,liang2020video,park2022per} are introduced, employing a memory network to extract vital information from a time-based buffer composed of all previous video sequences. 
Building upon this methodology, DPSTT~\cite{li2022rethinking} integrates a memory bank with decoupled transformers to track temporal lesion movement in medical ultrasound videos. 
However, DPSTT calls for substantial data augmentation to avoid overfitting and is marked by a sluggish processing speed, stressing some potential limitations. 
FLA-Net~\cite{flanet} presents a frequency and location feature aggregation network with a large amount of memory occupancy for ultrasound video breast lesion segmentation.
Thus, the challenge in medical video segmentation revolves around efficiently harnessing the wealth of temporal data available.

\subsection{State Space Models}

Recently, State Space Models (SSMs)~\cite{kalman1960new} have demonstrated notable efficiency in utilizing state space transformations~\cite{gu2021combining} to manage long-term dependencies within language sequences.
S4~\cite{gu2021efficiently} introduced a structured state-space sequence model to exploit long-range dependencies with the benefit of linear complexity.
Based on this, Mamba~\cite{gu2023mamba} integrates efficient hardware design and a selection mechanism employing parallel scan (S6), thereby surpassing Transformers in processing extensive natural language sequences.
Subsequently, S4ND~\cite{nguyen2022s4nd} explores SSMs' continuous-signal modeling of multi-dimensional data like images and videos.
More recently, Vision Mamba~\cite{zhu2024vision} and Vmamba~\cite{liu2024vmamba} pioneered generic vision tasks and outperformed transformer-based methods in effectiveness and efficiency, introducing bi-directional scan and cross-scan mechanisms to tackle the directional sensitivity challenge in SSMs.
U-Mamba~\cite{ma2024u} designed a hybrid CNN-SSM block, which is mainly composed of Mamba modules, to handle the long sequences in biomedical image segmentation tasks.
To the best of our knowledge, SSMs have not yet been explored in medical video segmentation tasks. 

\begin{figure*}[!t]
\centering
\includegraphics[width=0.9\textwidth]{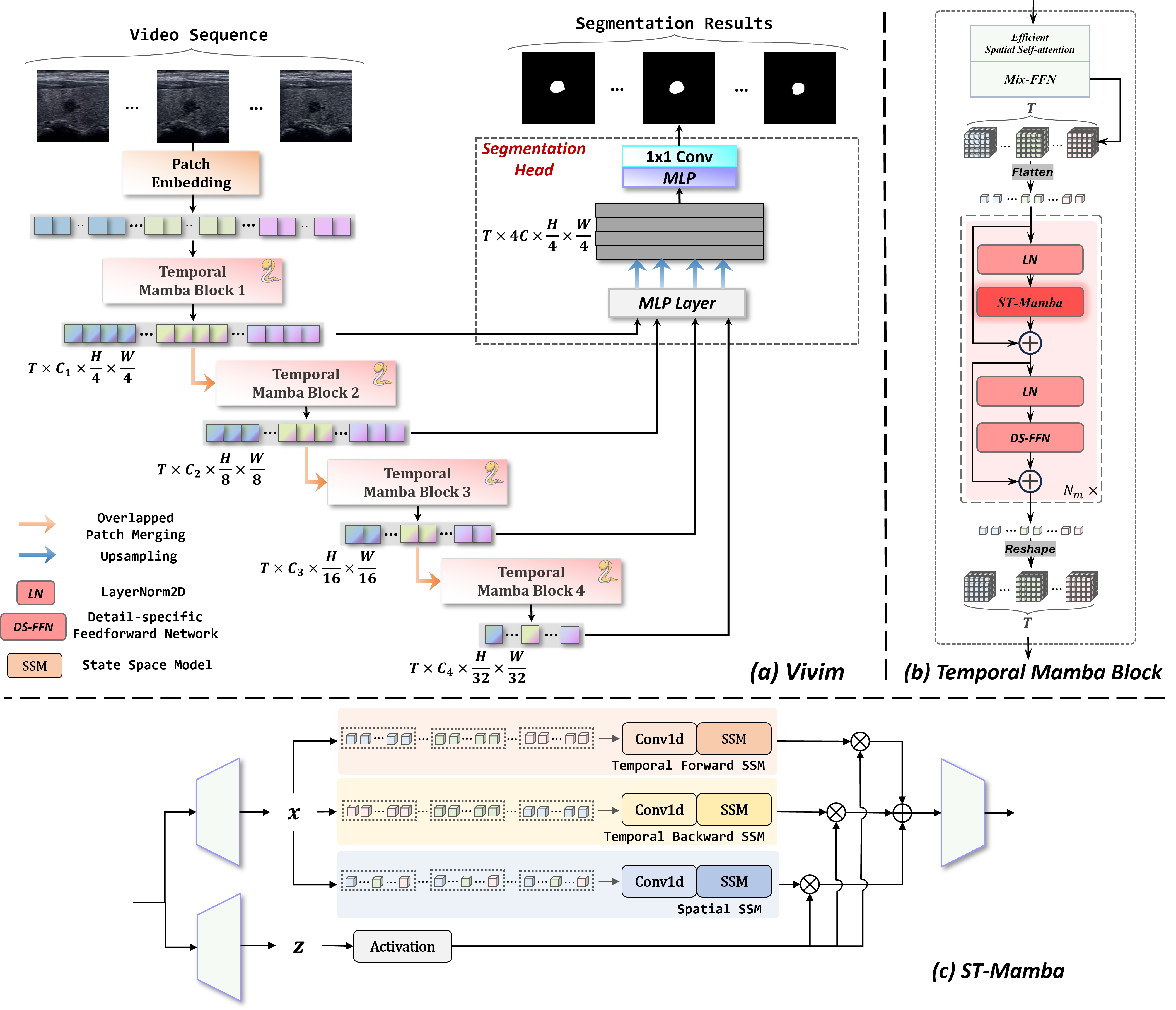}
\caption{
(a ) \textbf{The overview of the proposed Vivim for medical video segmentation.} The video sequence is first fed into patch embedding and multi-scale Temporal Mamba Blocks for encoding. Then, the feature sequences are aggregated to predict the segmentation results by a CNN-based segmentation head. (b) The fundamental building block of Vivim, namely Temporal Mamba Block. While Efficient Spatial Self-attention conducts initial spatial modeling, ST-Mamba explores spatiotemporal dependency in a linear complexity. (c) ST-Mamba incorporates spatiotemporal selective scan for long sequence modeling of video vision tasks in a multi-way spirit.
} 
\label{fig:method}
\end{figure*}

\section{Method}

\subsection{Overview}
In this part, we elaborate on a Mamba-based solution Vivim for medical video segmentation tasks. 
Our Vivim mainly consists of two modules: A hierarchical encoder with the stacked Temporal Mamba Blocks to extract coarse and fine feature sequences at different scales, and a lightweight CNN-based segmentation head to fuse multi-level feature sequences and predict segmentation masks.
Fig. \ref{fig:method} illustrates the flowchart of our proposed Vivim.
Specifically, given a video clip with $T$ frames, \ie, $\mathbf{V}=\{I^1,...,I^{T}\}$, we first divide these frames into patches of size $4\times4$ by overlapped patch embedding. We then feed the sequence of patches into our hierarchical Temporal Mamba Encoder to obtain multi-level spatiotemporal features with resolution $\{1/4, 1/8, 1/16, 1/32\}$ of the original frame. 
Finally, we pass multi-level features to the CNN-based segmentation head to predict the segmentation results.
The Boundary-aware Affine Constraint is deployed on the results only during training as shown in Fig.~\ref{fig:boundary}.
Please refer to the following sections for details of our proposed module.

\subsection{Preliminaries: State Space Models}
State Space Models (SSMs) are commonly considered as linear time-invariant systems, which map a 1-D function or sequence $x(t) \in \mathbb{R} \mapsto y(t) \in \mathbb{R}$ through a hidden state $h(t) \in \mathbb{R}^\mathtt{N}$. This system is typically formulated as linear ordinary differential equations (ODEs), which uses $\mathbf{A} \in \mathbb{R}^{\mathtt{N} \times \mathtt{N}}$ as the evolution parameter and $\mathbf{B} \in \mathbb{R}^{\mathtt{N} \times 1}$, $\mathbf{C} \in \mathbb{R}^{1 \times \mathtt{N}}$ as the projection parameters.
\begin{equation}
\label{eq:lti}
h'(t) = \mathbf{A}h(t) + \mathbf{B}x(t), \ 
y(t) = \mathbf{C}h(t).
\end{equation}

The discretization is introduced to primarily transform the ODE into a discrete function. This transformation is crucial to align the model with the sample rate of the underlying signal embodied in the input data, enabling computationally efficient operations.
The structured state space sequence models (S4) and Mamba are the classical discrete versions of the continuous system, which include a timescale parameter $\mathbf{\Delta}$ to transform the continuous parameters $\mathbf{A}$, $\mathbf{B}$ to discrete parameters $\mathbf{\overline{A}}$, $\mathbf{\overline{B}}$. The commonly used method for transformation is zero-order hold (ZOH), which is defined as follows:
\begin{equation}
\label{eq:zoh}
\mathbf{\overline{A}} = \exp{(\mathbf{\Delta}\mathbf{A})}, \ 
\mathbf{\overline{B}} = (\mathbf{\Delta} \mathbf{A})^{-1}(\exp{(\mathbf{\Delta} \mathbf{A})} - \mathbf{I}) \cdot \mathbf{\Delta} \mathbf{B}.
\end{equation}

After the discretization of $\mathbf{\overline{A}}$, $\mathbf{\overline{B}}$, the discretized version of Eq.~\eqref{eq:lti} can be rewritten as:
\begin{equation}
\label{eq:discrete_lti}
h_t = \mathbf{\overline{A}}h_{t-1} + \mathbf{\overline{B}}x_{t}, \ 
y_t = \mathbf{C}h_t.
\end{equation}

At last, the models compute output through a global convolution. 
\begin{equation}
\label{eq:conv}
\mathbf{\overline{K}} = (\mathbf{C}\mathbf{\overline{B}}, \mathbf{C}\mathbf{\overline{A}}\mathbf{\overline{B}}, \dots, \mathbf{C}\mathbf{\overline{A}}^{\mathtt{M}-1}\mathbf{\overline{B}}),\ 
\mathbf{y} = \mathbf{x} * \mathbf{\overline{K}},
\end{equation}
where $\mathtt{M}$ is the length of the input sequence $\mathbf{x}$, and $\overline{\mathbf{K}} \in \mathbb{R}^{\mathtt{M}}$ is a structured convolutional kernel.
\

\subsection{Overall Architecture}

\subsubsection{Hierarchical feature representation}
Multi-level features provide both high-resolution coarse features and low-resolution fine-grained features that significantly improve the segmentation results, especially for medical images.
To this end, unlike Vivit~\cite{arnab2021vivit}, our encoder extracts multi-level multi-scale features given input video frames.
Specifically, we perform patch merging frame-by-frame at the end of each Temporal Mamba Block, resulting in the $i$-th feature embedding $\mathcal{F}_i$ with a resolution of $\frac{H}{2^{i+1}}\times \frac{W}{2^{i+1}}$.

\subsubsection{Temporal Mamba Block}
Exploring temporal information is critically important for medical video segmentation by providing dynamic appearance and motion cues.
However, MSA in vanilla Transformer architectures has quadratic complexity concerning the number of tokens~\cite{vaswani2017attention}. This complexity is pertinent for long feature sequences from videos, as the number of tokens increases linearly with the number of input frames. 
Motivated by this, we develop a more efficient block, Temporal Mamba Block, to simultaneously exploit spatial and temporal information by structured state space sequence models.

As illustrated in Fig. \ref{fig:method} (b), in the Temporal Mamba Block, an efficient spatial-only self-attention module is first introduced to provide the initial aggregation of spatial information followed by a Mix-FeedForwoard layer. We leverage the sequence reduction process introduced in \cite{xie2021segformer,wang2021pyramid} to improve its efficiency.
For the $i$-level feature embedding $\mathcal{F}_i \in \mathbb{R}^{T \times C_i \times H\times W}$ of the given video clip, we transpose the channel and temporal dimension, and flatten the spatiotemporal feature embedding into 1D long sequence $h_i \in \mathbb{R}^{C_i \times THW}$.
% , enabling highly efficient sequential modeling with less inductive bias.
%
Then, the flattened sequence $h_i$ is fed into layers of a Spatio-Temporal Mamba module (ST-Mamba) and a Detail-Specific Feedforward (DSF).
The ST-Mamba module establishes the intra- and inter-frame long-range dependencies while the DSF preserves fine-grained details by a depth-wise convolution with a kernel size of $3\times 3\times 3$.
% Note that, after the DSF, the feature sequence is returned to the original shape by the inverse operation.
%
The procedure in the stacked Mamba Layer can be defined as, where $l\in [1, N_m]$:
\begin{equation}
\begin{aligned}
h^l=&\operatorname{ST-Mamba}\left(\operatorname{LN}\left(h^{l-1}\right)\right)+h^{l-1}, \\
&h^l=\operatorname{DSF}\left(\operatorname{LN}\left(h^l\right)\right)+h^l.
\end{aligned}
\end{equation}
% where $\phi$ denotes the transposition and flattening operation, $\phi^{-1}$ denotes its inverse operation, $l\in [1, N_m]$. 
Finally, we return the output feature sequence to the original shape and employ overlapped patch merging to down-sampling the feature embedding.

\subsubsection{Decoder}
To predict the segmentation mask from the multi-level feature embeddings, we introduce a CNN-based segmentation head.
While our hierarchical Temporal Mamba encoder has a large effective receptive field across spatial and temporal axes, the CNN-based segmentation head further refines the details of local regions.
To be specific, the multi-level features $\{\mathcal{F}_1,\mathcal{F}_2,\mathcal{F}_3,\mathcal{F}_4\}$ from the temporal mamba blocks are passed into an MLP layer to unify the channel dimension. These unified features are up-sampled to the same resolution and concatenated together.
Third, a MLP layer is adopted to fuse the concatenated features $\mathcal{F}$.
Finally, The fused feature goes through a $1\times1$ convolutional layer to predict the segmentation mask $\mathcal{M}$. The segmentation loss $\mathcal{L}_{seg}$ consisting of pixel-wise cross-entropy loss and IoU loss is applied during training.

\begin{figure}[!t]
\centering
\includegraphics[width=0.85\linewidth]{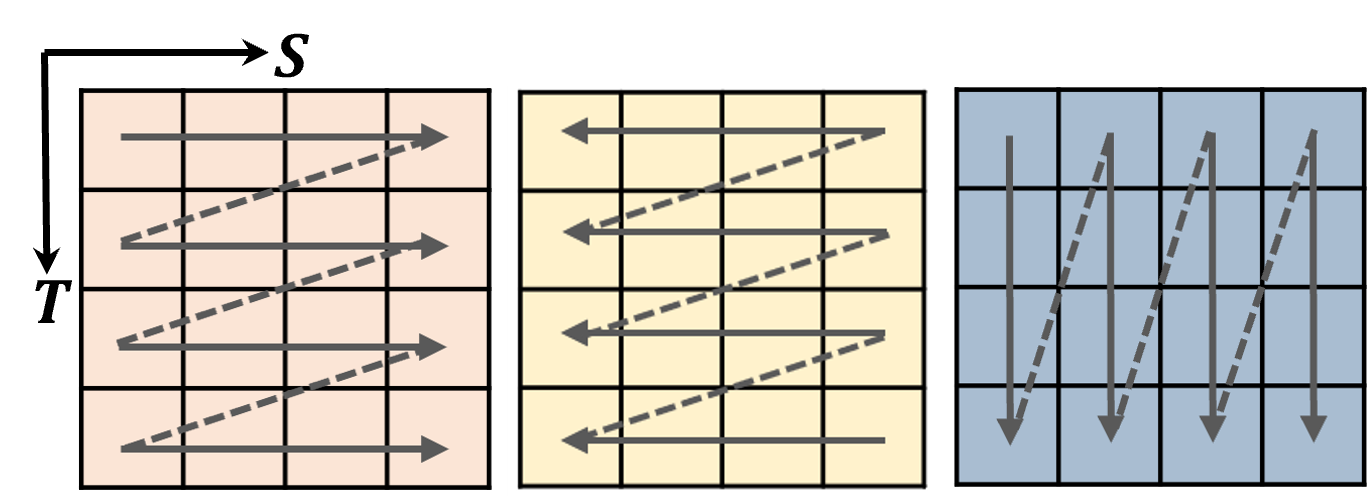}
\caption{
The illustration of the proposed spatiotemporal selective scan, including temporal forward scan, temporal backward scan and spatial scan.
} 
\label{fig:scan}
\end{figure}

\subsection{Spatiotemporal Selective Scan}

Despite the causal nature of S6 for temporal data, videos differ from texts in that they not only contain temporal redundant information but also accumulate non-causal 2D spatial information. 
To address this problem of adapting to non-causal data and fully exploring temporal information, we introduce ST-Mamba as shown in Fig.~\ref{fig:method} (c), which incorporates spatiotemporal sequence modeling for video vision tasks.

Specifically, to explicitly explore the relationship among frames, we first unfold patches of each frame along rows and columns into sequences, and then concatenate the frame sequences to constitute the temporal-first sequence $h_i^t\in \mathbb{R}^{C_i\times T(HW)}$. We parallelly proceed with scanning along the forward and backward directions to explore bidirectional temporal dependency.
This approach allows the models to compensate for each other's receptive fields without significantly increasing computational complexity.
Simultaneously, we stack patches along the temporal axis and construct the spatial-first sequence $h_i^s \in \mathbb{R}^{C_i\times (HW)T}$. We proceed with scanning to integrate information of each pixel from all frames.
The spatiotemporal selective scan mechanism with three directions is also vividly demonstrated in Fig.~\ref{fig:scan}.
Our mechanism explicitly considers both single-frame spatial coherence and cross-frame coherence, and leverages parallel SSMs to establish the intra- and inter-frame long-range dependencies. 
% resulting in more effective visual representation learning.
%
The structured state space sequence models with spatiotemporal selective scan (ST-Mamba), serve as the core element to construct the Temporal Mamba block, which constitutes the fundamental building block of Vivim.

\noindent\textbf{Computational-Efficiency. }
 SSMs in ST-Mamba and self-attention in Transformer both provide a crucial solution to model spatiotemporal context adaptively. Given a video visual sequence $\mathbf{K}\in \mathbb{R}^{1\times T\times M\times D}$, the computation complexity of a global self-attention and SSM are:
\begin{equation}
\label{eq:self-attn}
\Omega (\text{self-attention}) = 4\mathtt{(TM)}\mathtt{D}^2 + 2\mathtt{(TM)}^2\mathtt{D}, 
\end{equation}
\begin{equation}
\Omega (\text{SSM}) = 4\mathtt{(TM)}(2\mathtt{D})\mathtt{N} + \mathtt{(TM)}(2\mathtt{D})\mathtt{N}^2, 
\end{equation}
where the default expansion ratio is 2, $\mathtt{N}$ is a fixed parameter and set to 16. As observed, self-attention is quadratic to the whole video sequence length $\mathtt{(TM)}$, and SSM is linear to that. Such computational efficiency makes ST-Mamba a better solution for long-term video applications. This is also validated by the experimental analysis on the efficiency of ST-Mamba in Sec.~\ref{sec:efficiency}.

\begin{figure}[!t]
\centering
\includegraphics[width=\linewidth]{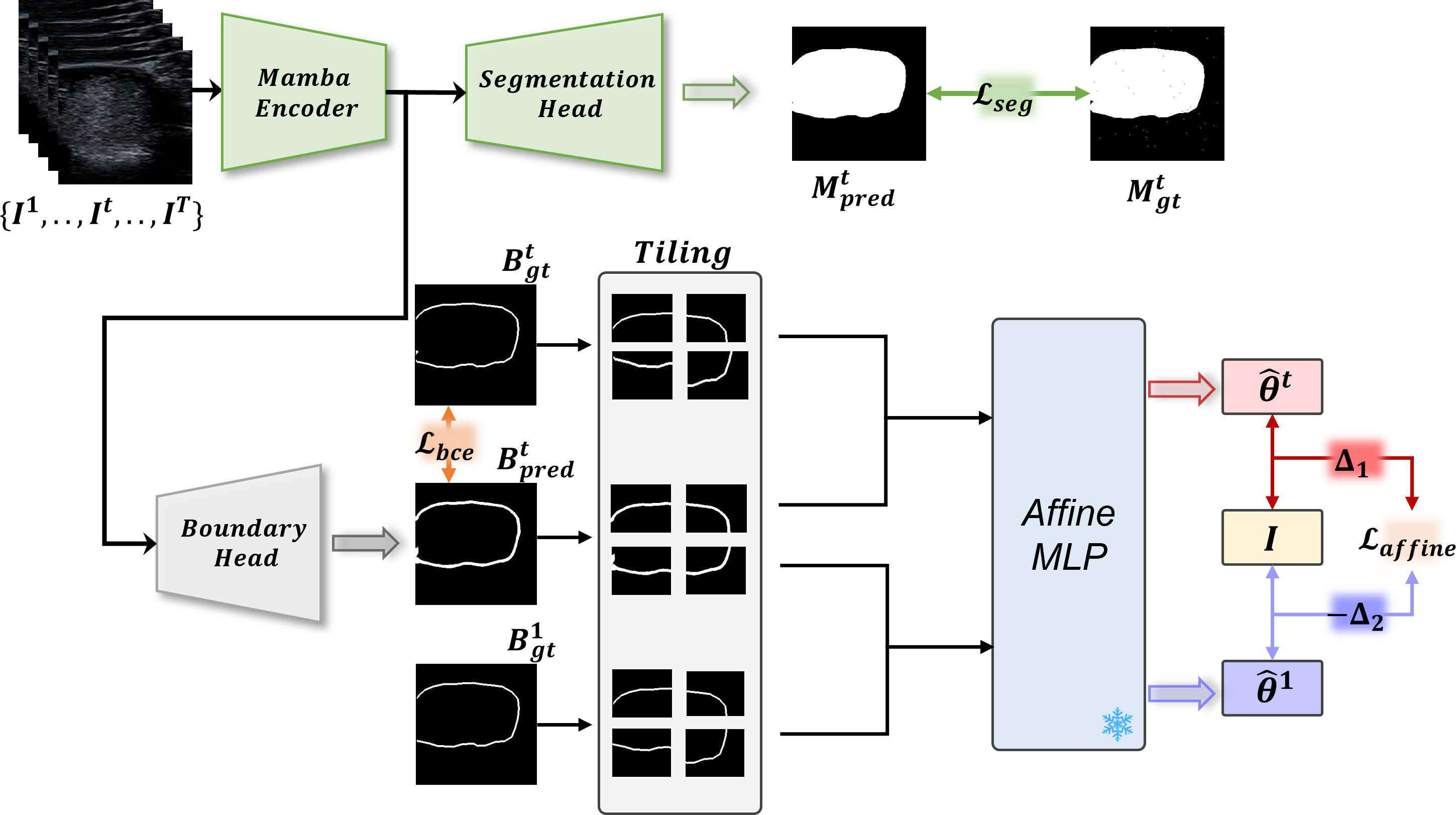}
\caption{
\textbf{The overview of the training strategy.} Specifically, our proposed patch-level boundary-aware affine constraint $\mathcal{L}_{affine}$ is introduced to optimize Vivim jointly with the segmentation loss $\mathcal{L}_{seg}$ and the boundary cross-entropy loss $\mathcal{L}_{bce}$. The pre-trained MLP for computing the affine transformation is frozen during training.
} 
\label{fig:boundary}
\end{figure}

\subsection{Boundary-aware Affine Constraint}
The network optimized only by the segmentation supervision tends to generate ambiguous and unstructured predictions, and overfit on training data. 
To mitigate these issues, we introduce a patch-level boundary-aware affine constraint inspired by InverseForm~\cite{borse2021inverseform} to enforce the predicted boundary structure.
Specifically, as illustrated in Fig.~\ref{fig:boundary}, we address this constraint task by optimizing the affine transformation between ground-truth boundaries and edges in feature maps towards identity transformation matrix. 
The ground truth edges within the patches are derived from applying the Sobel operator~\cite{canny1986computational} on ground truth masks, while an auxiliary boundary head consisting of three convolutional layers processes the feature patches from the Mamba encoder to obtain the predicted edge.
We calculate the affine transform matrix $\hat{\theta}_i^t$ for the $i$-th patch between ground-truth edge $B^t_{gt}$ and predicted edge $B^t_{pred}$ of the target frame $I^t$ in a video clip, by a pre-trained MLP.
Simultaneously, we calculate another affine transform matrix $\hat{\theta}_i^1$ for the $i$-th patch between ground-truth edge $B^1_{gt}$ of $I^1$ and predicted edge $B^t_{pred}$ of $I^t$ in a video clip.
This MLP is trained in advance with edge masks and not optimized during our method’s training. 
We optimize the matrix $\hat{\theta}_i^t$, and adversarially optimize $\hat{\theta}_i^1$ towards identity matrix $\mathbb{I}$ by:
\vspace{-2mm}
\begin{equation}
    \mathcal{L}_{affine} = \frac{1}{N_p}\sum_{i=1}^{N_p} \left(\Delta_1 \cdot \left|\hat{\theta}_i^t-\mathbb{I}\right|_F - \Delta_2 \cdot \left|\hat{\theta}_i^1-\mathbb{I}\right|_F \right),
    \label{eq:if}
\end{equation}
where $N_p$ denotes the number of patches and $\left|\cdot\right|_F$ is Frobenius norm. 
$\Delta_1$ and $\Delta_2$ is two balancing hyper-parameters to control the effects of $\hat{\theta}_i^t$ and $\hat{\theta}_i^1$, empirically set as 1.00 and 0.01.
In this objective, $B^t_{pred}$ is pushed toward $B^t_{gt}$ and pulled away from $B^1_{gt}$ to improve the target boundary and maintain the subtle inter-frame discrepancy in lesion structure.

% \vspace{-2mm}
We also employ the binary cross entropy loss $\mathcal{L}_{bce}$ between the whole predicted boundary and corresponding ground truths of the target frame to optimize the boundary detection further.
Finally, the overall loss to optimize during training is as follows, where the scaling parameters $\lambda_1, \lambda_2$ are both empirically set as 0.3:
% \vspace{-2mm}
\begin{equation}
    \mathcal{L}_{total} = \mathcal{L}_{seg} + \lambda_1\mathcal{L}_{affine} + \lambda_2\mathcal{L}_{bce}.
    \label{eq:overall_loss}
\end{equation}
% \vspace{2mm}

\begin{table*}[!t]
    \centering
    \caption{Quantitative comparison with state-of-the-art methods on our VTUS dataset (thyroid nodule) and the BUV2022 dataset (breast lesion). Dice, Jaccard, Precision and Recall are adopted as our evaluation metrics. The best scores are highlighted in bold.} 
    % \vspace{-2mm}
    \label{tab:brats23}
    \renewcommand\arraystretch{1.3}
    \resizebox{0.95\textwidth}{!}{
    \begin{tabular}{c | c| c| c |c |c |c | c| c| c| c |c }
    \hline
    \hline
    \multirow{2}{*}{Methods} & \multirow{2}{*}{Venue} &\multirow{2}{*}{Type} & \multicolumn{4}{c|}{\textbf{VTUS}} & \multicolumn{4}{c|}{\textbf{BUV2022}} & \multirow{2}{*}{FPS} \\
    \cline{4-11}
     &  & & Dice & Jaccard  & Precision  & Recall & Dice & Jaccard  & Precision  & Recall&  \\

    \hline
    
    UNet~\cite{ronneberger2015u} & MICCAI15 & image & 0.6662	&0.5328&	0.6703&	0.7471    & 0.7303  & 0.6247 & 0.7946  & 0.7272  & \textbf{88.18} \\
    UNet++~\cite{zhou2018unet++} & DLMIA18 & image & 0.7656&	0.6486&	0.7441&	0.8496   & 0.7179  & 0.6124 & 0.8280 &0.6884  &40.90\\
    TransUNet~\cite{chen2021transunet} & arXiv21& image & 0.7461&	0.6250&	0.7468&	0.8321	  &0.6547 & 0.5358  & 	0.7167   & 0.6682  &65.10  \\
    SETR~\cite{zheng2021rethinking} & CVPR21& image & 0.7288&	0.6010&	0.7399&	0.8089  & 0.6649  & 0.5480& 	0.7533   & 0.6643  &21.61 \\
    DAF~\cite{wang2018deep} & MICCAI18 & image & 0.7716&	0.6583	&0.7046	&0.8599	& 0.7890& 0.6954&0.7992 & 0.7979 &47.62 \\
        \hline
    OSVOS~\cite{perazzi2017learning} & CVPR17 & video & 0.7769	&0.6754	&0.7895&	0.8241	  & 0.7098 & 0.5674 & 	0.7778   & 0.6404  &27.25 \\
    ViViT~\cite{arnab2021vivit} & ICCV21 & video & 0.7610	&0.6459	&0.7789	&0.8252		& 0.6739 & 0.5446  & 	0.7554  & 0.6683  &24.33\\

    STM~\cite{oh2019video} & ICCV19 & video & 0.7898	&0.6897	&0.8112	&0.8251	 & 0.7862  & 0.6858& 	0.8201  &0.7910 &23.17  \\
    AFB-URR~\cite{liang2020video} & NIPS20 & video & 0.7930 & 0.6957 & 	0.7764 & 0.8429 & 0.8018   & 0.7034& 	0.8008 & 0.8591 	&11.84 \\
    DPSTT~\cite{li2022rethinking} & MICCAI22 & video & 0.8063&	0.7117	&0.8238	&0.8352	  & 0.8255& 0.7364  &\textbf{0.8389} & 0.8455  &30.50 \\
    FLA-Net~\cite{flanet} & MICCAI23 & video &0.8042 & 0.7075 &0.8121 & 0.8276  & 0.8232& 0.7315& 0.8334& 0.8422& 31.22\\
    RMem~\cite{zhou2024rmem} & CVPR24 & video &0.7804 & 0.6775 &0.7821 & 0.8298  & 0.7912& 0.6901& 0.8024& 0.8221& 29.54\\
    \hline
    \rowcolor{gray!15} 
    Our method  & -- -- & video & \textbf{0.8324} & \textbf{0.7391} & \textbf{0.8363} & \textbf{0.8711} & \textbf{0.8356} & \textbf{0.7450}  & 0.8357 & \textbf{0.8869} & 35.33 \\
    \hline
    \hline
    \end{tabular}
    }
    \label{tab:breast}
    % \vspace{-4mm}
\end{table*}

% \vspace{-2mm}
\section{Experiments}

% \vspace{-2mm}
\subsection{Dataset}
We evaluate our Vivim on three medical video segmentation tasks, \ie, video thyroid ultrasound segmentation, video breast lesion ultrasound segmentation and video polyp segmentation.

\begin{figure*}[!t]
\centering
\includegraphics[width=0.95\textwidth]{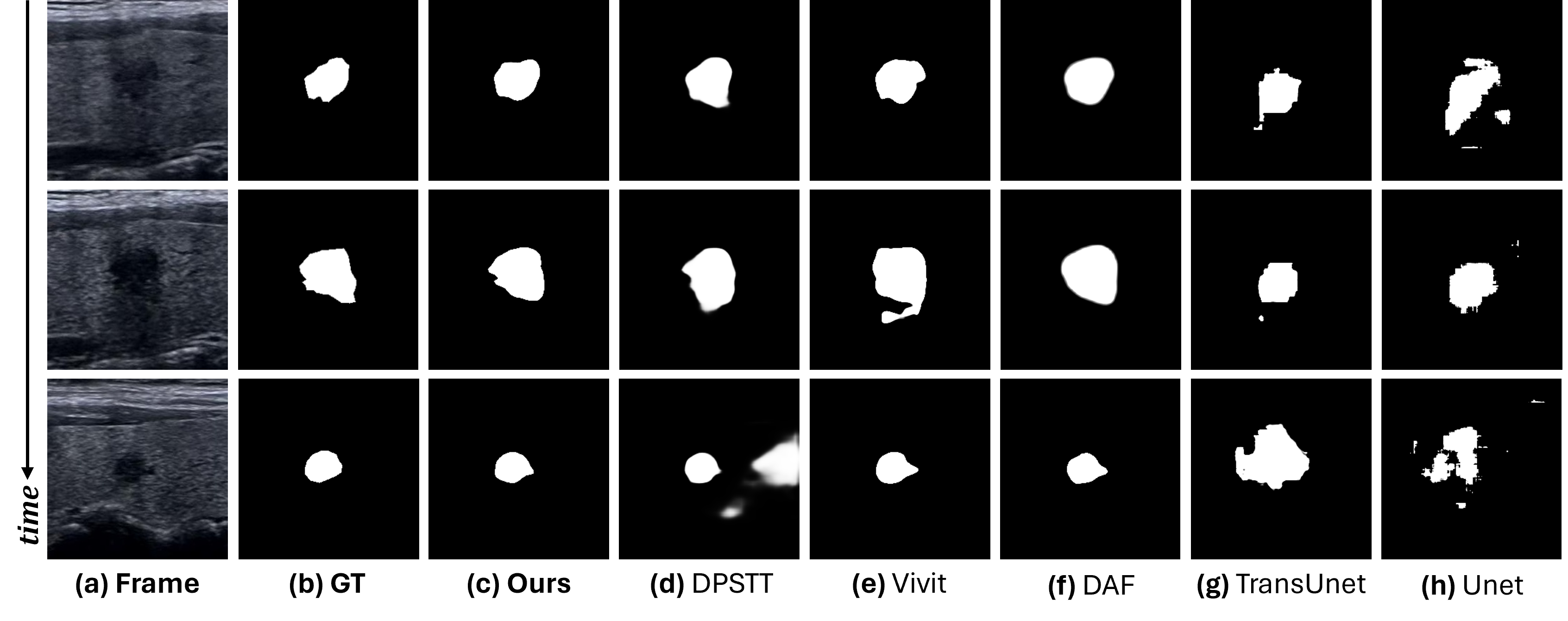}
\caption{
Visual comparison on video ultrasound thyroid segmentation with several competitive image- and video-based methods. Consecutive results of one case are displayed.
} 
% \vspace{-4mm}
\label{fig:vis_vtus}
\end{figure*}

\subsubsection{Video thyroid ultrasound segmentation}
We collect a video thyroid ultrasound segmentation dataset \textit{VTUS}. VTUS comprises 100 video sequences, one video sequence per patient, and a total of 9342 frames with pixel-level ground truth. 
VTUS contains the transverse viewed and the longitudinal viewed B-mode ultrasound videos captured by Mindray resona8/TOSHIBA Aplio500 vendors. These videos are cross-annotated by three experts with over three years of experience in thyroid diagnosis. The number of frames in these videos vary from 31 to 196 for better diversity.
The entire dataset is partitioned into training and test sets by 7:3, yielding a total of 70 training videos, 30 test videos.

\subsubsection{Video breast lesion ultrasound segmentation}
We conduct experiments on the BUV2022 dataset~\cite{li2022rethinking} consisting of 63 video sequences, with one video sequence per person, containing 4619 frames that have been annotated with pixel-level ground truth by experts. 
Following the approach outlined in~\cite{li2022rethinking}, the video sequences with spatial resolutions ranging from 580×600 to 600×800 were further cropped to a spatial resolution of 300×200.
We follow the official splits for training and testing.

\subsubsection{Video polyp segmentation}
We adopt four widely used polyp datasets, including image-based Kvasir~\cite{jha2020kvasir} and video-based CVC-300~\cite{bernal2012towards}, CVC-612~\cite{bernal2015wm} and ASU-Mayo~\cite{tajbakhsh2015automated}. Following the same protocol as \cite{ji2021progressively}, we train our model on Kvasir, ASU-Mayo and the training sets of CVC-300 and CVC-612, and conduct three experiments on test datasets CVC-300-TV, CVC-612-V and CVC-612-T.

\subsection{Implementation details}
The proposed framework was trained on one NVIDIA RTX 4090 GPU and implemented on the Pytorch platform.
Our framework is empirically trained for 100 epochs in an end-to-end way and the Adam optimizer is applied. The initial learning rate is set to $1 \times 10^{-4}$ and decayed to $1 \times 10^{-6}$. 
During training, we resize the video frames to $256 \times 256$, and feed a batch of 4 video clips, each of which has 5 frames, into the network for each iteration.
%

% \vspace{-2mm}
\subsection{Comparsion with Other Methods}

\subsubsection{Results on thyroid and breast lesion US video segmentation}
We employed four segmentation evaluation metrics, including Dice, Jaccard, Precision and Recall; for their precise definitions, please refer to \cite{wang2018deep}. We also report the inference speed performance by computing the number of frames per second (FPS).

% \noindent\textbf{Quantitative comparison. }
As shown in Tab.~\ref{tab:breast}, we quantitatively compare our method with many state-of-the-art methods on VTUS dataset and BUV2022 dataset. These methods including popular medical image segmentation methods (UNet~\cite{ronneberger2015u}, UNet++~\cite{zhou2018unet++}, TransUNet~\cite{chen2021transunet}, SETR~\cite{zheng2021rethinking}, DAF~\cite{wang2018deep}), and video object segmentation methods (OSVOS~\cite{perazzi2017learning}, ViViT~\cite{arnab2021vivit}, STM~\cite{oh2019video}, AFB-URR~\cite{liang2020video}, DPSTT~\cite{li2022rethinking}, FLA-Net~\cite{flanet}, RMem~\cite{zhou2024rmem}). For the fairness of comparisons, we reproduce these methods following their publicly available codes. 
Note that we adopted Vision Transformer as the backbone of FLA-Net.
We can observe that video-based methods tend to outperform image-based methods as evidenced by their better performance. This suggests that the exploration of temporal information offers significant advantages for segmenting thyroid nodules and breast lesions in ultrasound videos. 
More importantly, among all image-based and video-based segmentation methods, our Vivim has achieved the highest performance across all scores by a considerable margin
(\eg, 2.61\%, 2.74\% in Dice, Jaccard on VTUS, 1.01\%, 0.86\% in Dice, Jaccard on BUV2022 than the second-best method DPSTT).
Our Vivim also has the best run-time among all video-based methods observed from FPS.
This demonstrates that our solution can simultaneously learn spatial and temporal cues in an efficient way, and achieves significant improvements over those Transformer methods, such as SETR, ViViT and DPSTT. 
%
% \noindent\textbf{Qualitative comparison. }
As displayed in Fig.~\ref{fig:vis_vtus}, we visualize the thyroid segmentation results on the selected frames. Our model can better locate and segment the target lesions with more accurate boundaries.

\begin{table*}[t!]
  \centering
  \scriptsize
  \renewcommand{\arraystretch}{0.95}
  \setlength\tabcolsep{4.9pt}
  \caption{Quantitative results on three video polyp datasets. 
    The best scores are highlighted in \textbf{bold}.
    $\uparrow$ indicates the higher the score the better, and vice versa.
    }
    % \vspace{-2mm}
    \label{tab:polyp}
%   \vspace{-8pt}
\resizebox{0.9\textwidth}{!}{
  \begin{tabular}{rl||ccc|ccc|cc|c} 
  \hline
  \hline
  % \cline{3-11}
  && UNet &UNet++ &ResUNet & ACSNet & PraNet & PNS-Net  & LDNet & FLA-Net & \textbf{Vivim} \\
  & Metrics & MICCAI~\cite{ronneberger2015u} & TMI~\cite{zhou2018unet++} & ISM~\cite{jha2019resunet++} & MICCAI~\cite{zhang2020adaptive} & MICCAI~\cite{fan2020pranet} & MICCAI~\cite{ji2021progressively} & MICCAI~\cite{ldnet} & MICCAI~\cite{flanet} & \textbf{(Ours)} \\
\hline
% \hline
\multirow{6}{*}{\begin{sideways}CVC-300-TV\end{sideways}} 
& maxDice$\uparrow$ & 0.639   & 0.649   & 0.535   & 0.738   & 0.739   & 0.840 &  0.835   & 0.874   & \textbf{0.901}  \\
& maxSpe$\uparrow$ & 0.963   & 0.944   & 0.852   & 0.987   & 0.993   & 0.996 &  0.994   &  0.996 & \textbf{0.997}  \\
& maxIoU$\uparrow$ & 0.525   & 0.539   & 0.412   & 0.632   & 0.645   & 0.745  &  0.741   &  0.789 & \textbf{0.831}  \\
& $S_\alpha\uparrow$ & 0.793   & 0.796   & 0.703   & 0.837   & 0.833 & 0.909 &  0.898   &  0.907 &\textbf{0.928}  \\
& $E_\phi\uparrow$ & 0.826   & 0.831   & 0.718   & 0.871   & 0.852   & 0.921&  0.910   &  \textbf{0.969} & 0.958 \\
& $MAE\downarrow$ & 0.027   & 0.024   & 0.052   & 0.016   & 0.016   & 0.013 &  0.015   &  0.010 & \textbf{0.008} \\
\hline
\multirow{6}{*}{\begin{sideways}CVC-612-V\end{sideways}} 
& maxDice$\uparrow$& 0.725   & 0.684   & 0.752   & 0.804   & 0.869   & 0.873 &   0.870   &   0.885 & \textbf{0.897}\\
& maxSpe$\uparrow$ & 0.971   & 0.952   & 0.939   & 0.929   & 0.983   & 0.991 &   0.987   &   0.992 & \textbf{0.996} \\
& maxIoU$\uparrow$& 0.610   & 0.570   & 0.648   & 0.712   & 0.799   & 0.800 &   0.799   &   0.814 & \textbf{0.829}\\
& $S_\alpha\uparrow$ & 0.826   & 0.805   & 0.829   & 0.847   & 0.915   & 0.923  &   0.918   &   0.920 & \textbf{0.940}\\
& $E_\phi\uparrow$ & 0.855   & 0.830   & 0.877   & 0.887   & 0.936   & 0.944  &   0.941   &   0.963 & \textbf{0.971}\\
& $MAE\downarrow$ & 0.023   & 0.025   & 0.023   & 0.054   & 0.013   & 0.012  &   0.013   &   0.012  & \textbf{0.010}\\
\hline
\multirow{6}{*}{\begin{sideways}CVC-612-T\end{sideways}} 
& maxDice$\uparrow$ & 0.729   & 0.740   & 0.617   & 0.782   & 0.852   & 0.860 &  0.857    &  0.861& \textbf{0.872}\\
& maxSpe$\uparrow$ & 0.971   & 0.975   & 0.950   & 0.975   & 0.986   & 0.992 &  0.988    &  0.993& \textbf{0.995}\\
& maxIoU$\uparrow$ & 0.635   & 0.635   & 0.514   & 0.700   & 0.786   & 0.795 &  0.791    &  0.795& \textbf{0.810}\\
& $S_\alpha\uparrow$ & 0.810   & 0.800   & 0.727   & 0.838   & 0.886   & 0.903 &  0.892    &  0.904& \textbf{0.915}\\
& $E_\phi\uparrow$ & 0.836   & 0.817   & 0.758   & 0.864   & 0.904   & 0.903 &  0.903    &  0.904 & \textbf{0.921}\\
& $MAE\downarrow$ & 0.058   & 0.059   & 0.084   & 0.053   & 0.038   & 0.038 &  0.037    &  0.036 & \textbf{0.033}\\
    \hline
    \hline
  \end{tabular}
  }
  % \vspace{-2mm}
\end{table*}

\begin{figure*}[!t]
\centering
\includegraphics[width=\textwidth]{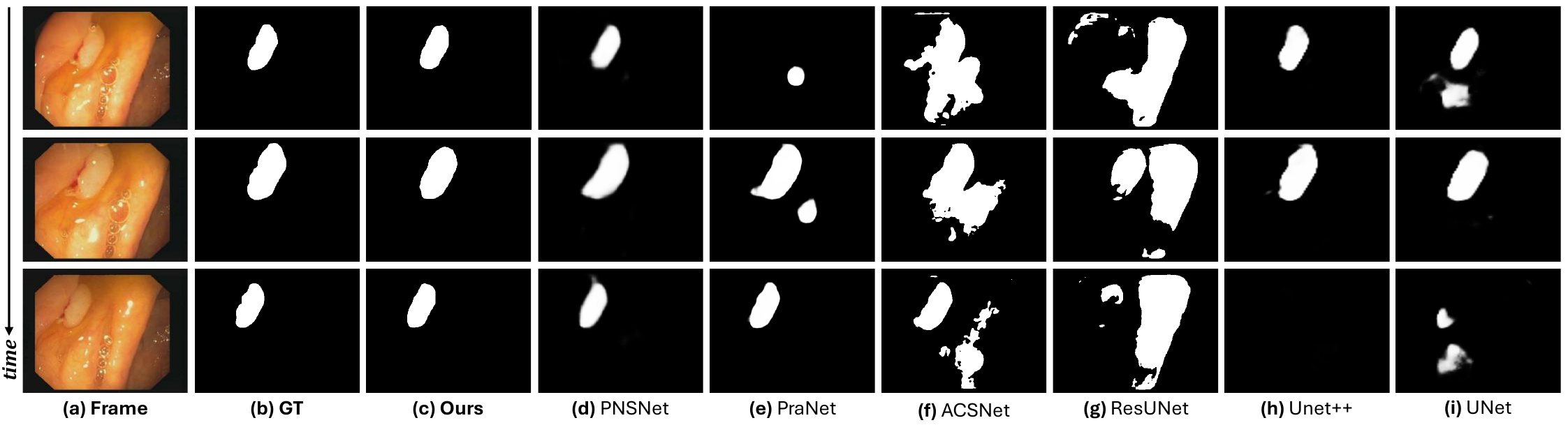}
% \vspace{-4mm}
\caption{Qualitative results on the selected frames of CVC-612-T. Our Vivim can better locate and segment polyps with more accurate boundaries than several competitive image- and video-based methods.}\label{fig:vis_polyp}
\end{figure*}

% \vspace{-4mm}

% \vspace{-2mm}
\subsubsection{Polyp video segmentation}
% \noindent\textbf{Metrics. }
We adopt six metrics following \cite{ji2021progressively}, \ie,  maximum Dice (maxDice), maximum specificity (maxSpe), maximum IoU (maxIoU), S-measure~\cite{fan2017structure} ($S_\alpha$), E-measure~\cite{fan2021cognitive} ($E_\phi$), and mean absolute error (MAE).

% \noindent\textbf{Quantitative comparison. }
We compare our method with existing methods as summarized in Tab.~\ref{tab:polyp}, including UNet~\cite{ronneberger2015u}, UNet++~\cite{zhou2018unet++}, ResUNet~\cite{jha2019resunet++}, ACSNet~\cite{zhang2020adaptive}, PraNet~\cite{fan2020pranet}, PNS-Net~\cite{ji2021progressively}, LDNet~\cite{ldnet} and FLA-Net~\cite{flanet}. We conduct three experiments on CVC-300-TV, CVC-612-V and CVC-612-T to validate the model’s performance. 
CVC-300-TV consists of both validation set and test set including six videos in total, while CVC-612-V and CVC-612-T each contain five videos. 
On CVC-300-TV, our Vivim achieves remarkable performance and outperforms all methods by a large margin (\eg, 2.7\% in maxDice, 2.2\% in maxIoU). On CVC-612-V and CVC-612-T, our Vivim consistently outperforms other SOTAs, especially 1.2\% and 1.1\% in maxDice, respectively.
% 
% \noindent\textbf{Qualitative comparison. }
We also visualize the polyp segmentation results on the consecutive frames of CVC-612-T in Fig.~\ref{fig:vis_polyp}. 
Our model demonstrates improved capability in locating and segmenting polyps with more precise boundaries.

\begin{table}[!t]
\centering
\caption{Ablation study of our Vivim design on VTUS dataset. In ST-Mamba, $T^f$ denotes temporal forward SSM, $T^b$ denotes temporal backward SSM, $S$ denotes spatial SSM, while BAC denotes boundary-aware affine constraint.
}
% \vspace{-2mm}
\small
\addtolength{\tabcolsep}{-1pt}
\resizebox{\linewidth}{!}{
\begin{tabular}{l|cccc|cccc} 
\toprule[0.15em]
\multicolumn{1}{c|}{\multirow{2}{*}{}} & \multicolumn{3}{c|}{\textit{ST-Mamba}}  & \multirow{2}{*}{\textit{BAC}} & \multicolumn{4}{c}{VTUS}   \\ 
\cline{2-4}\cline{6-9}
\multicolumn{1}{c|}{}                  & $T^f$ & $T^b$ & \multicolumn{1}{c|}{$S$} &                       & Dice$\uparrow$\  & Jaccard$\uparrow$\  & Precision$\uparrow$\  & Recall$\uparrow$\      \\ 
\hline
\hline
basic                                     &   -- &--    &--                        &--                         &     0.8144    &     0.7188   &  0.8040 &   0.8572    \\
C1                                     & \checkmark  &--    &--                        &--                         &    0.8159     &    0.7216    & 0.8170  &  0.8704   \\
C2                                     & \checkmark  & \checkmark  &--                        & --                        &     0.8213    &   0.7264     &  0.8239 &  0.8670     \\
C3                                     & \checkmark  & \checkmark  & \checkmark                      &  --                       &    0.8259     &   0.7310     &  0.8255 & \textbf{0.8753}    \\ 
\hline
\hline
\textbf{Ours}                                   & \checkmark  & \checkmark  & \checkmark                      & \checkmark                       &   \textbf{0.8324}      &    \textbf{0.7391}    &  \textbf{0.8363} & 0.8711    \\
\bottomrule[0.15em]
\end{tabular}
}
% \vspace{-4mm}
\label{tab:ablation}
\end{table}

% \vspace{-4mm}
\subsection{Ablation Study}
Extensive experiments are conducted on VTUS dataset to evaluate the effectiveness of our major components. To do so, we construct four baseline networks from our method.  
The first baseline (denoted as ``basic'') is to remove all Mamba layers and boundary-aware affine constraint from our network. It means that ``basic'' equals the vanilla SegFormer~\cite{xie2021segformer}.
Then, we introduce ST-Mamba layers with temporal forward SSM ($T^f$) into ``basic'' to construct another baseline network ``C1'', and further equip ST-Mamba with temporal backward SSM ($T^b$) to build a baseline network ``C2''. 
Based on ``C2'', spatial SSM ($S$) is incorporated into the ST-Mamba to construct ``C3''.
Hence, ``C3'' is equal to removing the boundary-aware affine constraint from the training of our network. 
Table~\ref{tab:ablation} reports the results of our method and four baseline networks. 
While ``basic'' performs competitively due to the pre-trained SegFormer weights on ADE20K, our proposed modules significantly advance its effectiveness.
Compared to ``basic'', ``C1'' has a great improvement across all metrics, which indicates that the vanilla SSM helps explore temporal dependency, thereby improving the segmentation performance in videos. 
Then, the better Dice and Jaccard results of ``C2'' over ``C1'' demonstrate that introducing our bidirectional temporal SSMs can critically benefit the cross-frame coherence.
Furthermore, by adapting SSMs to non-causal information, ``C3'' advances ``C2'' with a significant margin of 0.46\% in Dice and 0.83\% in Recall.
Finally, our method outperforms ``C3'' in terms of Dice, Jaccard and Precision, which indicates that the boundary-aware affine constraint can further help to enhance the thyroid segmentation results.

\begin{table*}[!t]
    \centering
    \caption{Ablation study for different attention modules. We feed a video clip of 32 frames with 256p into the three model variants and our method. ``TM'' denotes training memory, ``IM'' denotes inference memory, and ``OOM'' represents out-of-memory. ``Is Global'' describes whether the core modules are global modeling ones.} 
    % \vspace{-3mm}
    \label{tab:efficiency}
    \renewcommand\arraystretch{1.2}
    \setlength\tabcolsep{2pt}%调列距
    \resizebox{0.85\textwidth}{!}{
    \begin{tabular}{c| c| c c c c |c}
    \toprule[0.15em]
    \textbf{Methods} & \makecell{\textbf{Core Module}} & \makecell{\textbf{Input Size}}  & \makecell{\textbf{TM (M)}} & \makecell{\textbf{IM (M)}} & \makecell{\textbf{Run-time (s)}} & \textbf{Is Global} \\
    \hline

    M1  & \makecell{Spatio-temporal self-attention}  & $32\times 256^2$   & OOM  & -  & - & 	\Checkmark\\

    M2 & \makecell{Spatio-temporal Window self-attention}  & $32\times 256^2$   & 25,861  & 7,795  & 0.142 & \XSolidBrush \\

    M3  & \makecell{Spatio-temporal Factorized self-attention} & $32\times 256^2$   & 29,110  & 9,288  & 0.156 & 	\XSolidBrush\\

    \hline
    \rowcolor{gray!15} 
    \textbf{Our method} & Spatio-temporal Mamba & $32\times 256^2$  & \textbf{19,216} & \textbf{5,112} & \textbf{0.121} & \Checkmark \\
    
    \toprule[0.15em]
    % \vspace{-2mm}
    \end{tabular}
    }
\end{table*}

\begin{figure}[!t]
\centering
\includegraphics[width=\linewidth]{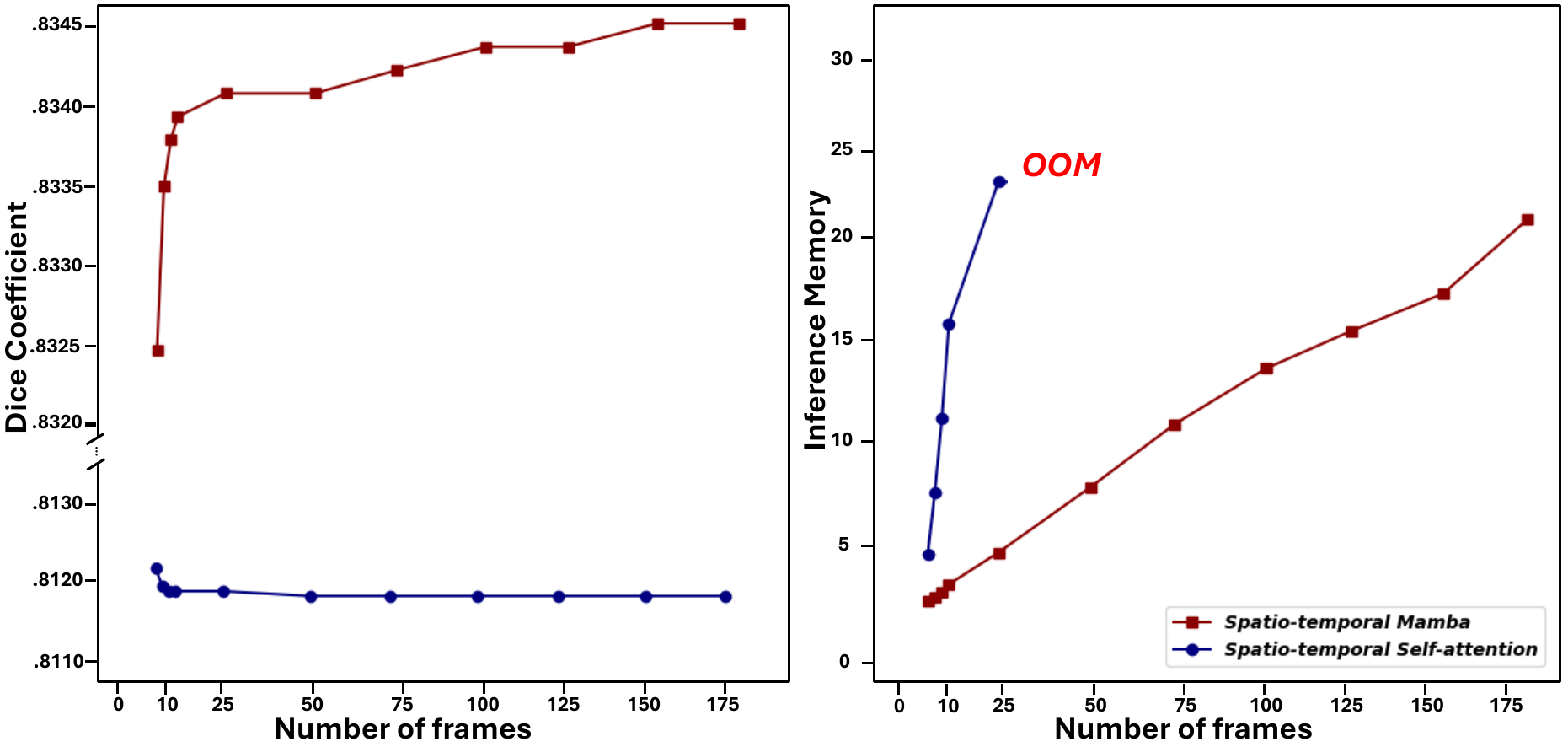}
\caption{
ST-Mamba performs better in effectiveness and efficiency when addressing long sequence modeling. (a) More reference frames can help improve the segmentation performance of ST-Mamba, but it is not applicable for spatio-temporal self-attention. (b) Vivim has a lighter memory burden than traditional attention-based methods when increasing the sequence length.
} 
\label{fig:efficiency}
\end{figure}

\subsection{Analysis on Efficiency of ST-Mamba}
\label{sec:efficiency}

We validate the high efficiency of the proposed ST-Mamba by two ablation studies presented in Tab.~\ref{tab:efficiency} and Fig.~\ref{fig:efficiency}. 
In Tab.~\ref{tab:efficiency}, we compare against several core modules for the modeling of spatio-temporal dependency, \ie, vanilla self-attention~\cite{vaswani2017attention}, window self-attention~\cite{liu2022video} and factorized self-attention~\cite{arnab2021vivit}. 
We replace ST-Mamba in our full Vivim with the three core modules to construct three variants M1, M2, and M3, respectively. 
We assessed the efficiency of these models using a single 48G A6000, considering Training Memory (TM), Inference Memory (IM), and Run-time as key metrics.
M1, incorporating 3D global self-attention to capture spatial and temporal information simultaneously, faces challenges due to the memory constraints when processing a video clip of 32 frames at a resolution of $256\times 256$. In contrast, M2 and M3 compromise on the receptive field to ensure that spatio-temporal modeling can be conducted within the available memory capacity.
Instead, our approach introduces an efficient global modeling module based on Mamba, leading to superior performance in terms of training memory, inference memory, and average run-time when compared to the other model variants.

Fig.~\ref{fig:efficiency} displays the Dice coefficient and memory costs with an increasing number of frames in one video clip at the inference stage. We evaluate M1 and our method, \ie, spatio-temporal self-attention and spatio-temporal Mamba, to verify the efficiency of our model.
As observed, M1 tends to maintain and even degrade the segmentation performance when referring to more neighboring frames. Instead, our method obtains an about 0.2\% improvement in Dice coefficient.
Furthermore, increasing the number of frames introduces an explosive growth in memory costs for M1. Our method can infer a video clip of over 150 frames using a single RTX4090.
This provides a great foundation for longer medical video segmentation within the limited memory capacity.

% \vspace{-4mm}
\section{Discussion}
Our Vivim is designed around the diagnostic process of radiologists that obtain the holistic understanding by ST-Mamba, and preserve local details by CNN-based decoder aggregating multiscale features and boundary constraint.
Although convolutional neural networks (CNNs) and Transformers have achieved impressive performance for many ultrasound video segmentation tasks~\cite{li2022rethinking,flanet}, there remains significant potential for enhancements in both efficiency and effectiveness.
A critical challenge limiting the broader application of CNNs and Transformers in medical video analysis is the trade-off between receptive field and computational complexity. 
This issue arises from the inherently local processing nature of CNNs and the high computational complexity associated with Transformers.
State Space Models (SSMs), \eg, Mamba, offer a more efficient technique for global dependency modeling compared to Transformers, facilitating more dynamic references within ultrasound videos. 
Ultrasound experts typically acquire the complete appearance of target tissues by utilizing both transverse and longitudinal views, which are captured by high-frame-rate devices. This leads to the requirement of long sequence modeling in ultrasound videos.
Unlike the self-attention mechanism in Transformers~\cite{vaswani2017attention,arnab2021vivit}, which scales quadratically with video sequence length, the computational complexity of SSMs scales linearly. 
This linear scalability makes SSMs well-suited for spatio-temporal joint modeling, allowing them to operate within the constraints of limited memory and computational resources, a feat challenging for Transformers, especially with longer video sequences.

The causal nature of SSMs is particularly well-aligned with tasks in Natural Language Processing (NLP) and video processing, where understanding the context in textual and temporal data is crucial~\cite{gu2023mamba}. A key challenge in adapting the Mamba model for medical video tasks lies in designing selective scan directions that effectively preserve non-causal spatial details of lesions and tissues while exploring temporal dependencies. 
Therefore, our approach goes beyond a straightforward application of Mamba; we establish a baseline that combines Transformers for spatial modeling with SSMs for spatio-temporal modeling. 
We introduce a tri-directional scan mechanism that simultaneously operates along temporal forward, temporal backward, and spatial forward directions, carefully balancing cross-frame coherence with single-frame spatial integrity.
In the future, our focus will be on further investigating a fully SSM-based solution for efficient medical video segmentation and developing more innovative selective scan mechanisms, tailored to medical video segmentation tasks for better lesion location. 
We will strive to better preserve the spatial correlation between neighboring patches of the target lesion when conducting 1-D sequence interaction.

\section{Conclusion}
In this paper, we present a Mamba-based framework Vivim to address the challenges of medical video segmentation, especially in modeling long-range temporal dependencies due to the inherent locality of CNNs and the high computational complexity of the self-attention mechanism. The main idea of Vivim is to introduce the structured state space models with spatiotemporal selective scan, ST-Mamba, into the standard hierarchical Transformer architecture. This facilitates the exploration of single-frame spatial coherence and cross-frame coherence in a computationally cheaper way than using the self-attention mechanism. An improved boundary-aware constraint at the training stage is proposed to mitigate the ambiguous prediction of our model.
We also contribute a video thyroid ultrasound segmentation dataset VTUS with 100 videos and 9342 annotated frames.
Experimental results on our collected VTUS dataset, ultrasound breast lesion videos and polyp colonoscopy videos reveal that Vivim outperforms state-of-the-art segmentation networks. Ablation studies also validate the superior efficiency of ST-Mamba to other spatio-temporal Transformer-based methods.

\bibliographystyle{IEEEtran}
\bibliography{refs}

\vfill

\end{document}